\documentclass[%
 aip,
 amsmath,amssymb,
 reprint,%
]{revtex4-1}

\usepackage{graphicx}
\usepackage{dcolumn}
\usepackage{bm}

\usepackage[utf8]{inputenc}
\usepackage[T1]{fontenc}
\usepackage{mathptmx}

\begin{document}

\preprint{AIP/123-QED}

\title{3D Convolutional Selective Autoencoder For Instability Detection in Combustion Systems}

\author{Tryambak Gangopadhyay}
\affiliation{Department of Mechanical Engineering, Iowa State University, Ames, IA, USA}

\author{Vikram Ramanan}
\affiliation{Indian Institute of Technology Madras, Chennai, India}

\author{Adedotun Akintayo}
\affiliation{Intel Corporation, Folsom, CA, USA}

\author{Paige K Boor}
\affiliation{Department of Mechanical Engineering, Iowa State University, Ames, IA, USA}

\author{Soumalya Sarkar}
\affiliation{Raytheon Technologies, East Hartford, CT, USA}

\author{Satyanarayanan R Chakravarthy}
\affiliation{Indian Institute of Technology Madras, Chennai, India}

\author{Soumik Sarkar}
\email{soumiks@iastate.edu}
\affiliation{Department of Mechanical Engineering, Iowa State University, Ames, IA, USA}




\begin{abstract}
While analytical solutions of critical (phase) transitions in physical systems are abundant for simple nonlinear systems, such analysis remains intractable for real-life dynamical systems. A key example of such a physical system is thermoacoustic instability in combustion, where prediction or early detection of an onset of instability is a hard technical challenge, which needs to be addressed to build safer and more energy-efficient gas turbine engines powering aerospace and energy industries.
The instabilities arising in combustion chambers of engines are mathematically too complex to model. This complexity is due to high-dimensional, pseudo-periodic coupling from high pressure oscillations and vorticities at the flame fronts generated by the combustion instabilities. To address this issue in a data-driven manner instead, we propose a novel deep learning architecture to detect the evolution of self-excited oscillations using spatiotemporal data, i.e., hi-speed videos taken from a swirl-stabilized combustor (a laboratory surrogate of gas turbine engine combustor). Our deep learning architecture called 3D convolutional selective autoencoder (3D-CSAE) consists of filters to learn, in a hierarchical fashion, the complex visual and dynamic features related to combustion instability from the training videos. We train the 3D-CSAE on frames of videos obtained from a limited set of operating conditions. We select the 3D-CSAE hyper-parameters that are effective for characterizing hierarchical and multiscale instability structure evolution by utilizing the dynamic information available in the video. The proposed model clearly shows performance improvement in detecting the precursors and the onset of instability compared to the ubiquitous 2D deep learning models, that lacks the ability to capture the temporal dynamics. The machine learning-driven results are verified with physics-based off-line measures. Advanced active control mechanisms can directly leverage the proposed online detection capability of 3D-CSAE to mitigate the adverse effects of combustion instabilities on the engine operating under various stringent requirements and conditions. 
\end{abstract}

\maketitle


\section{Introduction}
In many physical systems, critical transitions can occur in split seconds especially for combustion-dependent power generating systems. Hence, early detection of critical transitions is important for effective control of such physical systems exhibiting complex dynamics.
As the aerospace and energy industries are striving for more sustainable, energy efficient and low-NOx (nitrogen oxides) emitting gas turbine engines, they are adopting ultra-lean premixed and pre-vaporized combustors as the heart of energy production in an engine. Although ultra-lean combustion is environment-friendly, the operating regime of the fuel-air ratio is designed to move further from near-stoichiometric conditions and inch closer towards the safety margin for extracting every bit of efficiency out of the combustor. This particular design change makes the engine susceptible to an undesirable phenomenon called combustion instability. Combustion instability gives rise to serious anomalies such as lean blow out and rapid engine wear and tear due to increased vibration. Therefore, it is necessary to detect precursors of impending instability and mitigate its impact immediately for effective operation of gas turbines. Researchers have focused on the different aspects of combustion instability in these review articles \cite{ducruix2003combustion, candel2009flame, lieuwen2012unsteady, dowling2003acoustic}.

Combustion instability can be characterized by high amplitude flame oscillations at discrete tones or frequencies. Combustion instability, especially the thermo-acoustic variant is caused by a positive coupling between the heat release rate oscillations and the pressure oscillations and its resonant frequencies typically represent the natural acoustic modes of the combustor. The lower-order models, which rely on the role of velocity perturbations in interacting with the flame, have assumptions such as smooth cross-section and laminar flow characteristics \cite{dowling1995calculation,noiray2008unified,stow2009time}. However, in actual combustors operating at very high Reynolds number and employing flame holders, along with low-frequency chaotic pressure tones, the combustor flow exhibits a recurring spatiotemporal activity called coherent structure during instability. Coherent structures are defined as fluid mechanical structures associated with coherent phase of vorticity \cite{hussain1983coherent}, which sheds at the duct acoustic modes forced by high pressure amplitudes causing large scale oscillations in flow velocity and overall flame shape by curling and stretching.
The interaction of coherent structures with the flame is still not well understood with the study of generation (and persistence) of such structures in non-reacting turbulent flows at a nascent stage. 

\begin{figure*}
\centering
\includegraphics[width=12cm, keepaspectratio]{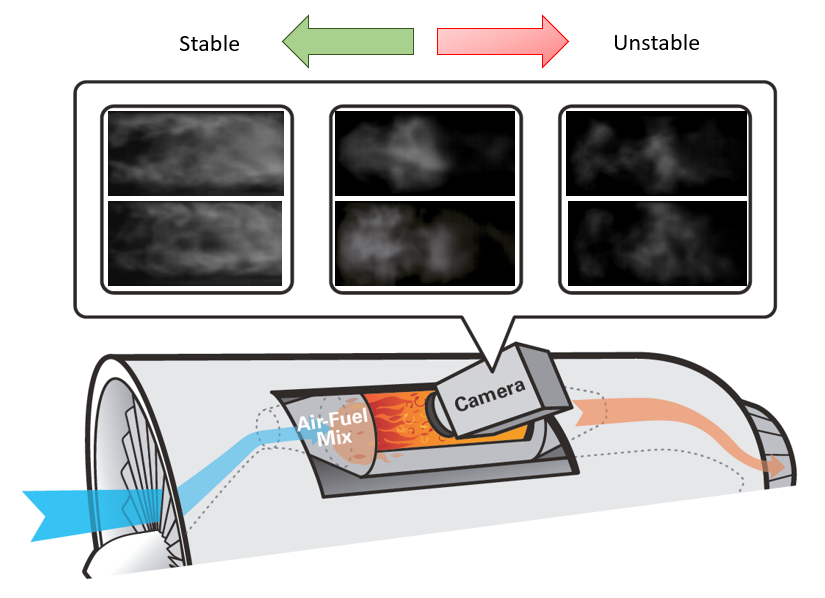}
\caption{Illustration of the concept of capturing hi-speed flame videos in an engine combustion chamber; example images from the laboratory scale combustor are shown for stable, relatively stable and unstable conditions (two images each). }
\label{engine_figure} 
\end{figure*}

Over the past few decades, researchers have attempted to understand and build detectors for combustion instability by physics-based modeling, data-driven algorithms or hybrid approaches. Majority of the efforts has been spent in characterizing the combustor pressure signal and detecting temporal precursors by various reduced order modeling, time series analysis \cite{gotoda2011dynamic,murayama2018characterization}, and signal processing \cite{nair2014multifractality,sen2018dynamic,sen2016investigation,barwey2019experimental} or machine learning approaches \cite{sarkar2015early}. 
Lately it has been observed that flame images or videos can reveal important information related to coherent structure dynamics that may help in identifying precursors with more lead time compared to only pressure signal analysis \cite {sarkar2015early1}. The common methods for coherent structure detection are rotational symmetry \cite{ramanan2019investigation}, proper orthogonal decomposition (POD) \cite{berkooz1993proper,locurto2018mode} (similar to principal component analysis\cite{bishop2006pattern}) and dynamic mode decomposition (DMD) \cite{schmid2010dynamic,ghosal2016detection}, which utilize spectral theory to derive spatial coherent structure modes. Although these methods have widespread applications in understanding spatiotemporal flow characteristics, it is shown that they demonstrate high parametric sensitivity for instability detection purposes and their applicability does not extend to noisy line-of-sight videos at per with more realistic data collection scenario. Significant studies in instabilities of swirl flames have also been conducted in the framework of flame transfer functions in the context of frequency-amplitude dependence and the underlying flow physics \cite{kim2013generalization,palies2011modeling,bellows2007flame,thumuluru2009characterization,palies2011nonlinear}. In particular, flame transfer functions provide low-order model-based tools that have been used to predict instability frequency and amplitude by solving the nonlinear dispersion relations \cite{noiray2008unified,cosic2014nonlinear,han2015prediction}.

Using artificial neural networks and clustering methods, studies \cite{barwey2019experimental,barwey2019using,barwey2020extracting} have been performed in the domain of combustion systems, especially on fluid mechanical aspects of combustion and identifying mapping between flow-field and heat release rate markers. 
Researchers have also shown early instability detection capability by combining complex networks and machine learning \cite{kobayashi2019early} and have used Bayesian machine learning for combustion diagnostics \cite{sengupta2020bayesian}.
Recently, the thermo-fluids research community has explored deep learning for various surrogate modeling and prediction problems~\cite{ling_kurzawski_templeton_2016,de2019deep,murata2020nonlinear,deng2019cnn,pawar2019deep,raissi2020hidden,kutz2017deep,fukami2020assessment}.
In the present context, deep learning models have been developed \cite{gangopadhyay2018characterizing,gangopadhyay2019deep,gangopadhyay2020deep} to classify conditions based on combustion instability, but there has been no rigorous validation of the results using physics-based principles. Also, these models rely only on classifying the conditions but not on evolution of the instability structures in the flames, which make these models ineffective to detect transitions.
2D autoencoder based deep learning frameworks \cite{akintayo2016prognostics, han2020combustion} can extract flame structures but neglect the inherent temporal dynamics of the instability phenomenon.

\begin{figure*}
\centering
\includegraphics[width=17cm, keepaspectratio]{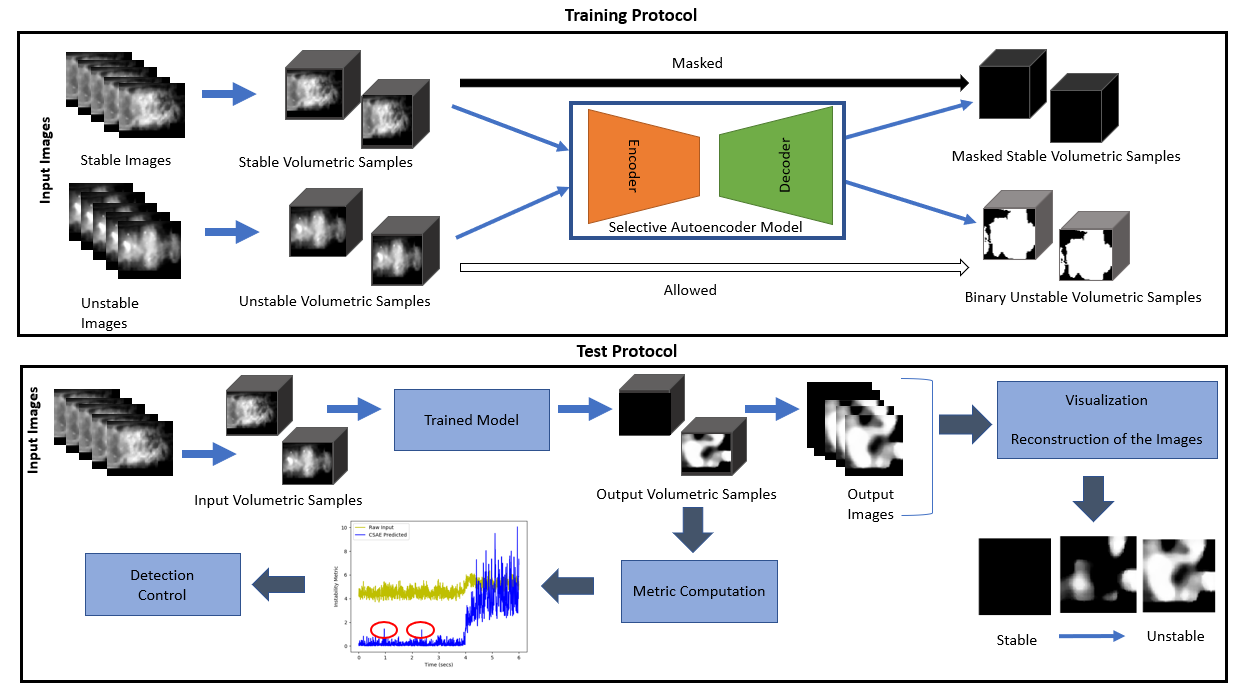}
\caption{Overall framework with the training and test protocols. In the training process, the stable volumetric samples are completely masked whereas the unstable volumetric samples are fully reconstructed (in a binarized form - with zero intensity as threshold), thus making the autoencoder learn in a selective manner. The quasi-static training datasets for stable and unstable conditions are determined by the domain experts. For testing, separate datasets are used demonstrating transition from stable to unstable regimes. The trained model computes the output volumetric samples and a nonzero output suggests the appearance of instability-related features (e.g., coherent structures) in the flame images. Instability metric (based on K-L divergence) distribution computed from the output samples shows a prominent transition across time and also identifies early warning signals.}
\label{framework} 
\end{figure*}

Although there exists a significant number of approaches to detect instability early in a combustor, they lack few necessary properties in part such as robustness to a broad range of operating regimes, speed of detection by ingesting real data and detection accuracy for active control purposes. Also, most of them have not yet explored the recent advancements in deep learning to extract adequate spatiotemporal information from the combustion flames. In this context, we propose a novel deep learning architecture called 3D convolutional selective autoencoder (3D-CSAE) that is able to detect precursors and onset of combustion instability from flame video snippets. As it is difficult to reliably collect data for instability precursors or onset of instability transitions, we train such a model with completely stable and completely unstable flame data. The primary learning objective is to be able to completely mask the stable flame signatures while reliably reconstructing unstable flame videos. With this training objective, 3D-CSAE can capture time evolution of flame coherent structures, which in turn enables the model to detect any instability signatures well before complete transition to unstable region. We train the 3D-CSAE on hi-speed videos taken from a swirl-stabilized combustor which is a laboratory-scale representation of gas turbine engine combustor for a limited set of operating conditions. The model is still shown to be effective for a diverse set of operating conditions, demonstrating the generalizability of such learning process. The performance of 3D-CSAE is also significantly better than its 2D deep learning counterpart 2D-CSAE \cite{akintayo2016prognostics} which demonstrates the need for taking the temporal dynamics into account. As a 3D-CSAE based framework can perform online detection of precursors and onset of combustion instability, we envision that such a framework will enable active control mechanisms for mitigation of combustion instability.

\section{Methods}
In this section, we describe our data analytics and learning framework for processing the flame video snippets. The details of the dataset preparation are presented along with our proposed 3D-CSAE model. 

\subsection{Processing of 3D frames}
Multiple images are uniformly stacked to compute a single volumetric sample as shown in Fig.~\ref{framework}, with similar processing schemes implemented for the training and test protocols. 
From a typical hi-speed flame video $\textbf{V}$, sequential frames are extracted, where set $ \textbf{V} = \{ I_1, \cdots, I_t, \cdots, I_T \} $. Each $I_t \in \mathcal{R}^{H \times W}$ is an image with height $H$ and width $W$. The total number of frames in the sequence is $T$. For each image, after extracting the region of interest, it is resized to resolution 64 x 64. A 2D-CSAE consists of 2D convolution operations and therefore, 2D-CSAE takes individual frames as input. On the other hand, a 3D-CSAE is designed to capture the implicit temporal correlations in the sequential images. The input of the 3D-CSAE is a volumetric sample which is obtained by temporal stacking of the frames. 

Fig.~\ref{framework} shows examples of stable and unstable images being converted to stable and unstable volumetric samples for input to the 3D-CSAE model. 
The number of images in a volumetric sample is denoted by $N$. 
We parameterize the local dependency of the temporal frames by $N$ to experiment on multiple values yielding the most feasible number of sequential frames to be stacked. 
The resulting volumetric samples are therefore $ V^{(1i)} = \{ I_{1}, \cdots, I_{N} \}, V^{(2i)}  = \{ I_{1+k}, \cdots, I_{N+k} \}, \cdots,  V^{(ji)}  = \{ I_{1+(j-1)k}, \cdots, I_{N+(j-1)k} \}, \cdots $. 
The stacked volumetric samples are indicated by the index $j$. The temporal gap between the volumetric samples is denoted by $k$. With a dataset sufficient enough to train our model, there was no need for data augmentation. Therefore, by setting $k = N$, we ensure there is no overlap of frames between two successive volumetric samples.

\subsection{Dataset}
\begin{figure*}
\centering
\includegraphics[width=16cm, keepaspectratio]{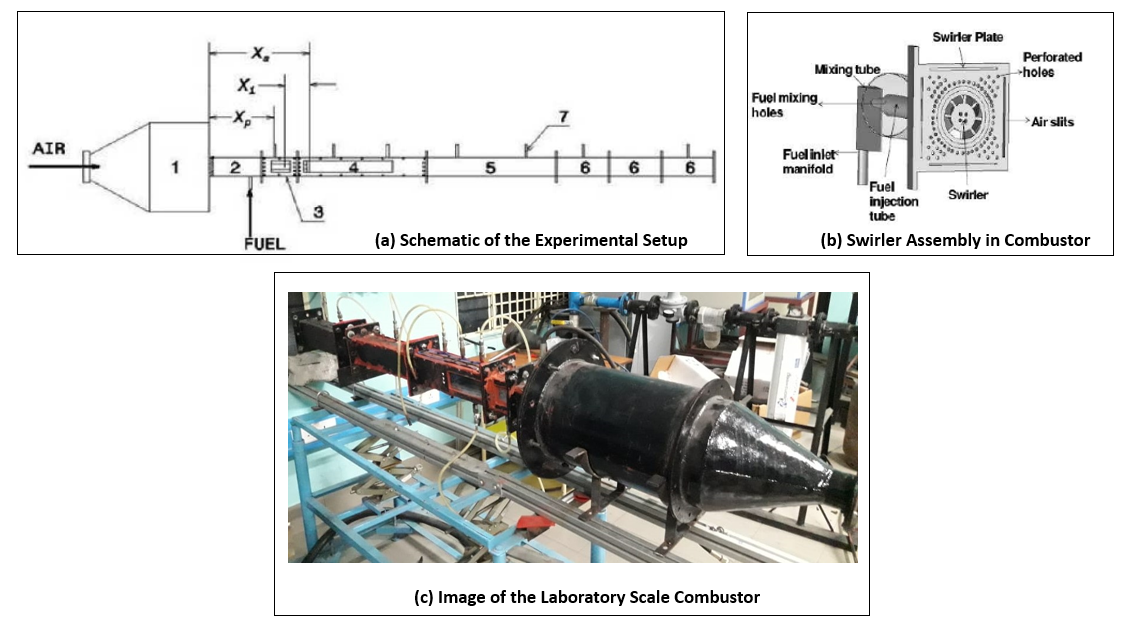}
\caption{The laboratory-scale combustor used for data collection. (a) Schematic of the experimental setup. 1 - settling chamber, 2 - inlet duct, 3 - IOAM, 4 - test section, 5 - big extension duct, 6 – small extension ducts, 7 - pressure transducers, Xs - swirler location measured downstream from settling chamber exit, Xp - transducer port location measured downstream from settling chamber exit, Xi - fuel injection location measured upstream from swirler exit, (b) Swirler assembly used in the combustor.}
\label{experimental_setup} 
\end{figure*}

To train our model, the ground truth labels (stable, unstable) for the image sequences are provided by the domain experts. 
For dataset collection, we induce combustion instability in a laboratory-scale swirl combustor (Fig.~\ref{experimental_setup}), which has a swirler of diameter 30 mm and vane angles of 60 degrees. Air is provided to the combustor through a settling chamber of diameter 28 cm and thereafter, through a square cross section of side 6 cm. The experimental setup includes an inlet section, an inlet optical access module (IOAM), a primary combustion chamber and a secondary duct. The IOAM facilitates optical access to the fuel tube. The fuel is injected co-axially with air through a fuel injection tube, having slots on the surface at selected distances upstream of the swirler. 

For the training dataset, the chosen upstream distances are 90 mm and 120 mm. For the upstream distance of 90 mm, there occurs partial premixing of the fuel with air, while the distance of 120 mm facilitates full premixing of the fuel and air. Based on the swirler diameter, two inlet Reynolds numbers (Re) are chosen for a fixed fuel flow rate of 0.495 g/s. The lower Re (7,971) and higher Re (15,942) exhibit stable and unstable conditions respectively. In another way, the inlet Re is kept fixed at 10,628 and two different fuel flow rates are used. The fuel flow rates 0.308 g/s and 0.66 g/s exhibit unstable and stable behaviours respectively. The hi-speed images are captured at 3000 Hz (with resolution of 1024 x 1024) for 3 secs at each of these conditions. Therefore, for the training dataset, the ground truth labels (stable, unstable) for the flame images are provided by the domain experts. After data cleaning, we have around 35,000 frames each for stable and unstable condition in the training dataset. 


The test dataset is collected separately with the focus on whether our trained model can capture if a transition occurs from stable to unstable state. 
With the sampling frequency of 3000 Hz, a 6 second test video comprises of 18,000 flame images, which are then converted to a sequence of $18000/N$ volumetric samples.
For processing the test dataset, we use the same $N$ and $k$ values as that of the training dataset.
If $N$ is not exactly a factor of 18000, then the required number of frames are omitted from the end. 
We identify the test protocols by the ratio of air flow rate (AFR) and fuel flow rate (FFR). Both AFR and FFR are expressed in lpm (liters per minute). 

\begin{enumerate}
	\item $500_{40to30}$: Protocol has AFR = 500 lpm and FFR is varied from 40 lpm to 30 lpm.
	\item $600_{50to35}$: Protocol has AFR = 600 lpm and FFR is varied from 50 lpm to 35 lpm.  
	\item $500to600_{40}$: Protocol has FFR = 40 lpm and AFR is varied from 500 lpm to 600 lpm. 	  
\end{enumerate} 

We present the detailed results for the protocol $500_{40to30}$ in the main content. For the protocols $600_{50to35}$ and \\ $500to600_{40}$, more detailed results are shown in the supplementary material.

\subsection{Instability Metric}
We propose a metric based on the Kullback Leibler (KL) divergence \cite{kullback1951information} to get an estimation of instability in the volumetric samples. 
The extent of instability of a frame is measured from the expected reconstruction of a stable frame, which is complete masking. 
KL divergence $D(I)$ is computed for each frame $I$, which can be expressed as follows:

\begin{equation}
D(I) = \sum_{i \in I} lim_{R(i)\to0+} I(i)\log \frac {I(i)} {R(i)}
\end{equation}

Each pixel in the frame $I$ is represented as $i$. $R$ is the reference image, having pixel values almost equal to zero. Higher value of D(I) indicates higher deviation from R (expected stable reconstruction) and therefore, it is a sign of more instability. 
The average value of D(I) from the $N$ images is considered to be the instability metric of a volumetric sample. 
This metric can be used as a common measure of instability for all the test protocols. 

\subsection{3D Convolutional Selective Autoencoder (3D-CSAE)}
Autoencoders, which can learn meaningful representations without any requirement for labels, fall among the self-supervised learning techniques. 
In an autoencoder, a compression function compresses the input information and a decompression function reconstructs the input from the compressed representation. 
The encoder and decoder are representatives of the compression and decompression functions in an autoencoder model. The dimension reduction in the encoder is very important as a low dimensional embedding captures a noise-less representation of the input and leads to better learning of the salient features. But, more compression in the encoder stage can lead to greater information loss as well. From the low dimensional embedding, the decoder network reconstructs a high dimensional representation. The weights of an autoencoder model are learned automatically from the training samples and no explicit annotation is required. The parameters of the encoder and decoder are optimized to minimize the reconstruction loss. 

\begin{figure*}
\centering
\includegraphics[width=16cm, keepaspectratio]{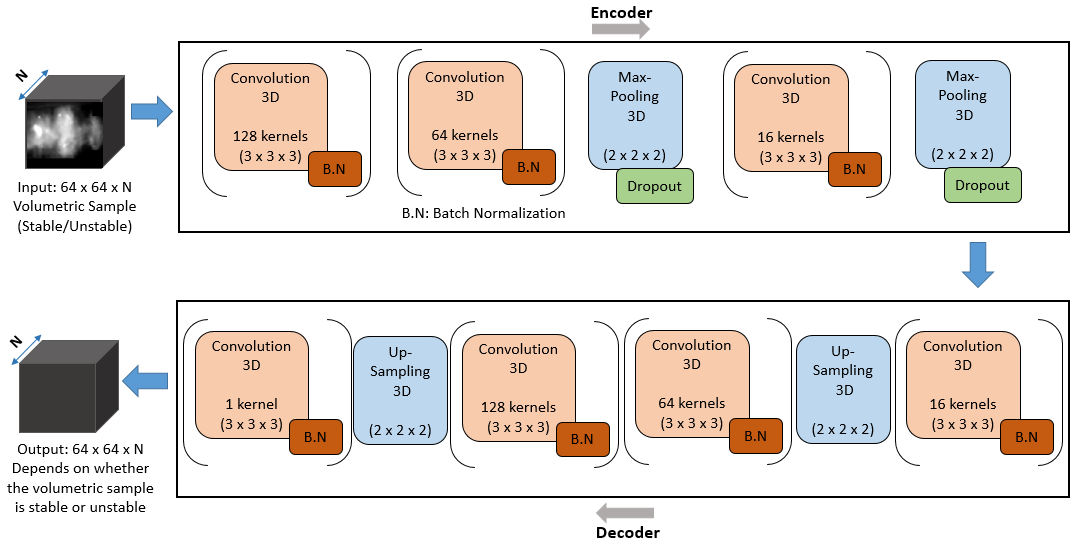}
\caption{The details of the 3D-CSAE Model. The model takes in input a volumetric sample comprising a sequence of consecutive flame images. The output reconstruction, which is also a volumetric sample, depends on whether the input represents stable or unstable regime. The 3D-CSAE model comprises of 3D convolution, 3D max-pooling and 3D up-sampling layers.}
\label{3dcsae_model} 
\end{figure*}

When autoencoders are built for images, convolutional neural networks are used as encoders and decoders. 
2D convolutional autoencoder approach has been previously applied successfully for applications in the domain of computer vision \cite {lore2017llnet, pu2016variational}. 
In this paper, we propose 3D-CSAE, a 3D convolutional autoencoder model, in which the encoder takes in volumetric samples as input and computes an informative low dimensional representation which acts as input for the decoder part. The model is trained to learn selectively such that stable data samples are completely masked whereas the unstable data samples are reconstructed as reliably as possible. Note that here we consider a binarized reconstruction with zero intensity as the threshold. The training protocol of 3D-CSAE is shown in Fig.~\ref{framework}. 

Each volumetric sample comprises of $N$ frames as described above. 
There is a limitation on the value of $N$, as stacking multiple frames can be a limitation on the basis of the GPU memory. We experiment with different values of $N$ and observe the performance on the validation set. We find that $N$=16 is an appropriate number of frames to construct the volumetric data samples. To finalize the 3D-CSAE architecture (Fig.~\ref{3dcsae_model}), we search for the best set of other model hyper-parameters through trial and error experiments, especially keeping in consideration the quality of dimension reduction in the encoder and output reconstructions for the validation set. 

In 3D-CSAE, the convolution layers are initialized with a number of uniformly random 3D kernels (filters). The batches of volumetric samples when scanned through these layers compute the feature maps by performing the convolution operation. The features are transformed by a rectified linear unit (ReLU) activation \cite{glorot2011deep}.  The layered architecture ensures that the convolution operation is performed at multiple abstraction levels of the volumetric samples. In the encoder, along with 3D convolution, 3D pooling layers are used which reduce the dimension of the feature maps. In the decoder, for reverting the sample to its original dimensions, 3D upsampling layers are used with more filters learned in the subsequent convolution layers. The number of trainable parameters for 3D-CSAE is about 0.51 million. For comparative study, we use 2D version of the model termed as 2D-CSAE. The architecture of 2D-CSAE is demonstrated in the supplementary material. 

In the 3D conv layers, the feature maps are computed in the form of 3-dimensional volumes to extract spatiotemporal information from the data. We use kernels with the dimensions of 3 x 3 x 3. The conv layers are initialized with a number of uniformly random 3D kernels. These scan through batches of volumetric frames' inputs. For the conv layers, zero padding is used to keep the input and output sizes same. The hyper-parameters related to each of these layers are the kernel size and the number of kernels.
During the training process, as the parameters of the layers are updated, the distribution of activations in each layer can change due to update of the parameters in the previous layers. This problem (called, the internal covariate shift) can be avoided with batch normalization \cite{ioffe2015batch}, which we employ after each 3D conv layer. Nonlinearity is imposed by using the rectified linear unit (ReLU) function.
The pooling layers reduce the dimensions of the feature maps and extract the important features. In the encoder of 3D-CSAE, we use two 3D max-pooling layers with the pool size of 2 x 2 x 2. Dropout used after the max-pooling layers help in reducing over-fitting. In the decoder, we use two 3D up-sampling layers, which perform the opposite function as that of the pooling layers. The up-sampling layers upscale the feature maps by repeating the rows and columns around the axes of symmetry.   

All the input features are normalized to have the range (0, 1). To learn the model parameters, we utilize the reconstructed output, given the selective training criteria of masking the stable volumetric samples and fully reconstructing the (binarized) unstable volumetric samples. The error in masking (or, not) of the appropriate regions are back-propagated to modify the model parameters (i.e., filter weights and biases). Mean squared error (MSE) is used as the loss function. To reduce the error, the parameters are adjusted in the optimization process. The dataset is randomly split into training (80\%) and validation (20\%) sets. The models are trained using NVIDIA GPUs for 200 epochs. After experiments with different optimizers and related parameters, Adam optimizer \cite{kingma2014adam} is finally used with a learning rate of 0.001 and momentum of 0.975.

\section{Results}
In this section, we demonstrate the performance of our proposed framework in detecting combustion instabilities. We also validate the outcomes of our data-driven approach using physics-based techniques.  

\subsection{Detection Using Deep Learning Model}
Upon training the CSAE model as described above, volumetric test data samples (i.e., flame video snippets) are fed into the trained model to get the output reconstructions. Visualizations of the reconstructions for the stable regime, during transition and unstable regime can be analyzed considering the selective training criteria. We compute the instability metric for both the input and output volumetric samples. The details of the test protocol is shown in Fig.~\ref{framework}.

\begin{figure*}
\centering
\includegraphics[width=17cm, keepaspectratio]{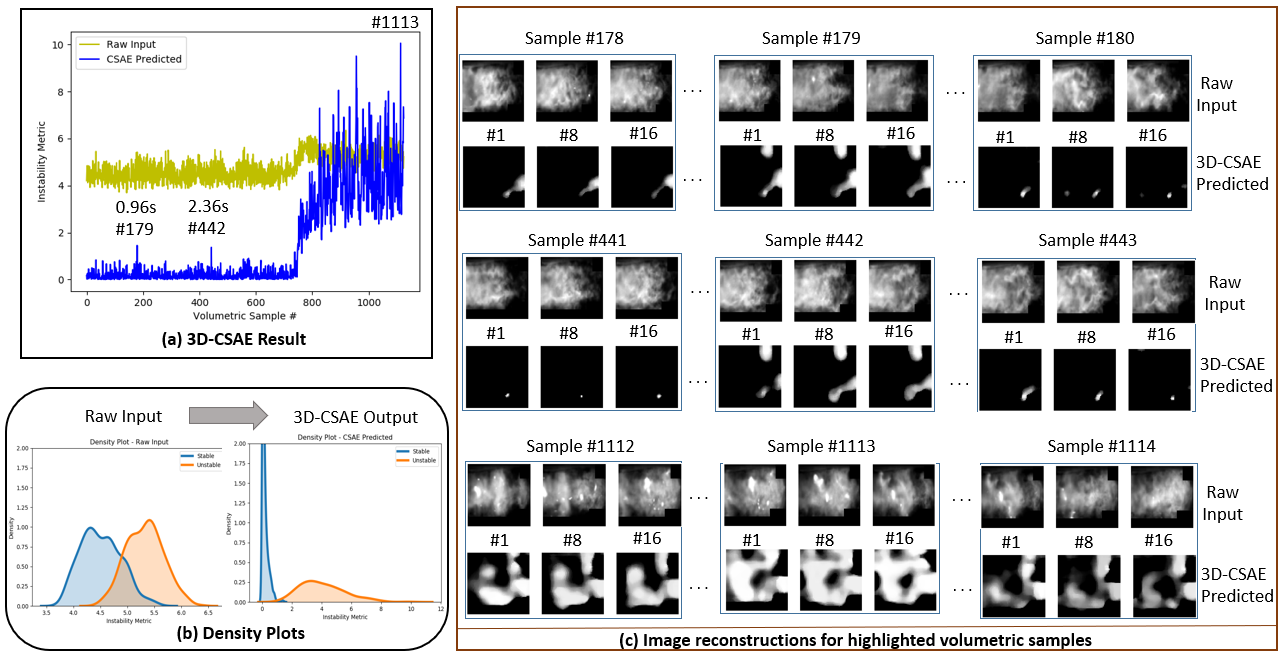}
\caption{The figure showing the detection results using 3D-CSAE for the test protocol $500_{40to30}$. (a) Instability metric computed on the CSAE predicted samples identifies a clearer transition compared to that on the raw input. 
(b) We observe that the overlap between the instability metric distributions for the stable and unstable samples diminishes in the 3D-CSAE output.
(c) The reconstructions for the highlighted volumetric samples are demonstrated. 
For each volumetric sample (having 16 frames), the reconstructions are shown for frames 1, 8 and 16.
}
\label{3dcsae_500_40_30_results} 
\end{figure*}

The results for the test protocol $500_{40to30}$ is shown in Fig.~\ref{3dcsae_500_40_30_results}. Each input volumetric sample is formed after stacking 16 consecutive frames and after, traversing the total duration of 6 secs (18,000 frames) in that manner, 1125 (18000/16) volumetric samples are generated. This input sequence of volumetric samples is referred to as the "Raw Input" in Fig.~\ref{3dcsae_500_40_30_results}(a). The corresponding output sequence of the samples reconstructed by 3D-CSAE is termed as "CSAE Predicted".
To reduce the noise slightly, we use image-wise mean filtering with a filter size of 6 on the 3D-CSAE output samples.
For each volumetric sample, instability metric is computed frame wise and then the average value is considered to be the sample wise metric. 
 
From the plots of the instability metric, we observe a much steeper transition for the 3D-CSAE predictions compared to the raw input, which highlights the discrimination capability of 3D-CSAE in distinguishing between features from stable and unstable data. 
To explain this further, we show the distributions of the instability metric values for stable and unstable samples in Fig.~\ref{3dcsae_500_40_30_results}(b). 
We use the density plot which is a smoothed, continuous version of histogram estimated using kernel density estimation.
A continuous curve (Gaussian kernel) is drawn at every individual data point and all the curves are added together to make a single smooth density estimation.
Therefore, the y-axis in the plot represents the probability density function for the kernel density estimation.
Volumetric samples before the transition at 4s (the transition time evident from the physics-based measure normalized conditioned pressure plot in Fig.~\ref{3dcsae_500_40_30_validation}(b) explained later) are considered as stable and those after that, as unstable. 
We observe that for the raw input, there is considerable overlap between the stable and unstable distributions. Thus even an optimal Bayesian decision boundary \cite{bishop2006pattern} will incur a significant error in classifying the stable and unstable regimes. However, for the 3D-CSAE output, this overlap diminishes and the distributions are well separated.

\begin{figure*}
\centering
\includegraphics[width=16cm, keepaspectratio]{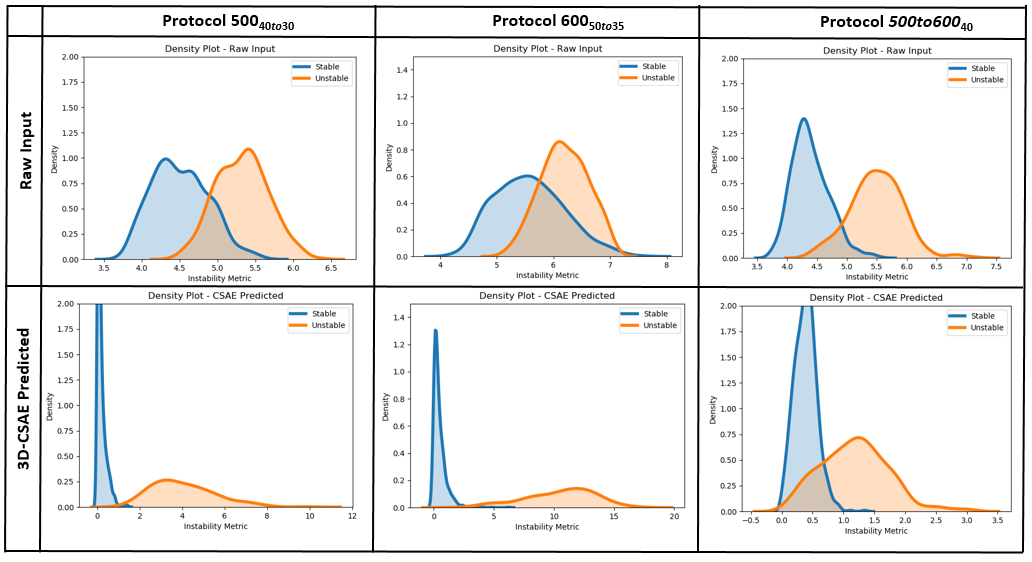}
\caption{The instability metric distribution plots for the raw input and the 3D-CSAE reconstructions under the three test protocols. For all the protocols, significant overlap between stable and unstable distributions are present based on the raw input data. Through extraction of most informative low dimensional features for masking the stable and reconstructing the unstable frames, 3D-CSAE distinguishes the distributions by reducing the overlap for the test protocols $500_{40to30}$ and $600_{50to35}$ - the protocols in which actual transition to instability is occurring. It implies that the 3D-CSAE is able to extract discriminating features from the stable and unstable data for these two protocols. But for the protocol $500to600_{40}$, the distribution overlap remains significant even in the 3D-CSAE outputs which reinforces the earlier interpretation that no proper transition to instability has occurred here.}
\label{density_plots} 
\end{figure*}

Density function plots of protocol $600_{50to35}$ from  Fig.~\ref{density_plots} further confirms our understanding that 3D-CSAE is discriminating the stable and unstable features. From the instability metric distribution plots in Fig.~\ref{density_plots}, the difference of the situations is also obvious. Typically for the protocols in which actual transition is happening (i.e., for the protocols $500_{40to30}$ and $600_{50to35}$), we see that the stable and unstable distributions become well separated when the data gets processed through the 3D CSAE model. However, the overlap between these distributions remain significant even after 3D CSAE processing for the protocol $500to600_{40}$ where proper transition to instability doesn't occur. Detailed results for the protocols $600_{50to35}$ and $500to600_{40}$ are shown in the supplementary materials.

As observed in Fig.~\ref{3dcsae_500_40_30_results}(a), the instability metric curve based on the raw input sequence does not show a considerable change to capture the transition. Also, there seems to be no clear indication or precursor prior to the transition while computing the instability metric using the raw input. 
In contrast, the transition from stable to unstable regime is much more prominent in the CSAE predicted sequence. Also, we observe two peaks for samples \#179 and \#442 in the stable regime, which can be perceived as indications that the system is getting closer to a transition (i.e., precursors of instability). To examine further, we first show the image reconstructions around the two samples \#179 and \#442. The frames in these samples show reconstruction of the flame structures, with less masking compared to the samples just before or after it. This happens as the trained model identifies image features similar to those in unstable frames in these supposed `stable frames' and hence, attempts to reconstruct them instead of masking them. From domain knowledge, it is known that certain coherent structures can indeed start forming at a smaller scale in stable flames prior to a transition \cite{hussain1983coherent, sarkar2015early1}. Therefore, our framework can be potentially used to generate scientific insights regarding the evolution of the coherent structures and their use as precursors to oncoming instability. 

We demonstrate the reconstructions around a volumetric sample (\#1113), a peak which falls in the unstable regime. From Fig.~\ref{3dcsae_500_40_30_results}(c), we can see that as the system enters the unstable regime, there is increased amount of flame structures (i.e., unstable flame features) and the masking effect reduces as evident from the frame reconstructions of samples \#1112, \#1113 and \#1114. 
This volumetric reconstruction capability gives 3D-CSAE a significant advantage compared to classification models based on convolutional neural network (CNN) \cite{gangopadhyay2018characterizing} and long short-term memory (LSTM) \cite{gangopadhyay2020deep} in the context of this problem. A 2D CNN based classification model does not consider the inherent temporal dynamics of this problem. While 2D CNNs coupled with LSTM can handle a temporal sequence of frames to classify the conditions based on instability \cite{gangopadhyay2019deep, gangopadhyayexplainable}, these classification models do not provide visual reconstructions of their decisions. In contrast, our proposed 3D-CSAE framework attempts to reconstruct spatiotemporal features related to instability that can be useful for both decision-making as well as scientific insights. A 2D counterpart of our 3D-CSAE framework can also provide similar outcomes while not considering the temporal dynamics. In the next section (and the supplementary material), we categorically demonstrate that it is critical to consider the dynamics by comparing the performance of the two variants of the CSAE approach. 
The results from the 3D-CSAE are further verified using off-line physics-based techniques, which we describe in the following section.  

\subsection{Validation Through Physics-Based Techniques}
In the previous section, we have showed that our proposed 3D-CSAE based framework can detect potential precursors or early warning signals well ahead of the transition to full-blown instability. To validate this claim, that is to show that these early detections are indeed related to instability and not due to noise or other data artifacts, we analyze our observations using off-line, post hoc physics-based techniques. Specifically, in this exercise, we use two well-known concepts called flame edge and normalized conditioned pressure for evaluation of flame edge.

\begin{figure*}
\centering
\includegraphics[width=18cm, keepaspectratio]{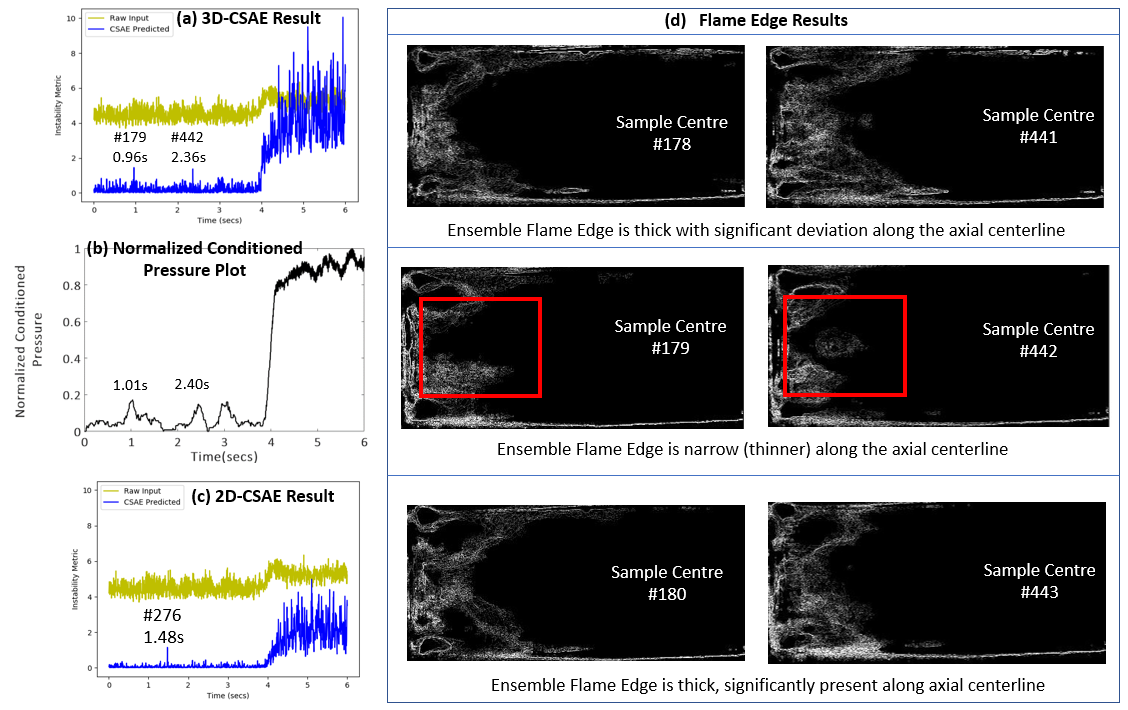}
\caption{The 3D-CSAE detection results for the test protocol $500_{40to30}$ are validated using physics-based approaches (normalized conditioned pressure and ensemble flame edge). Comparative results from 2D-CSAE are also shown. 
(a) From the 3D-CSAE results, the peaks are observed before the transition at 0.96s (sample \#179) and 2.36s (sample \#442).  
(b) Similar close peaks are observed from normalized  pressure plot, which also shows that 3D-CSAE can capture the transition correctly around 4 secs. 
(c) The 2D-CSAE fails to correctly identify the early-warning signals before the transition occurs, as evident from the highlighted peak.
(d) From the ensemble flame edge results, for samples \#179 and \#442, features (thin ensemble flame edge) are observed which are characteristic of instability.}
\label{3dcsae_500_40_30_validation} 
\end{figure*}

Flame edge is the interface between the burning and unburnt reactants \cite{turns1996introduction, taamallah2015thermo, sampath2016investigation}.
In the context of laminar flames with negligible unsteadiness, the flame edge is a time-stationary surface. However, for flames in turbulent flows, the instantaneous flame edge is not stationary due to underlying interactions between flow-field and flame. During unstable flame-acoustic interactions, it is known that the flame is largely modulated by ordered or coherent flow structures that have a length scale, which is significantly larger than most turbulence scales, especially at low frequencies, as in the present study. In such scenario, this effect blurs the effect of turbulence in inducing flame wrinkling. The instantaneous flame edge can be modulated due to the presence of both turbulent eddies (not periodic in a cumulative sense) and coherent structures (possessing definite frequency). The relative difference between the modulations is determined by the response of the flame to both modulation sources. 
Flame edge modulated by turbulent eddies are comparatively bigger in scale in comparison to that modulated by large coherent structures, especially at frequencies of interest in the present work. 

We formulate a physics-based validation technique based on the difference in orderliness observed during stable and unstable conditions.
Higher temporal and spatial orderliness is observed during unstable conditions, modulated by coherent structures. This orderliness is the classical signature of combustion instability driven by coherent structures. 
On the other hand, such orderliness is not demonstrated during stable conditions, modulated only by turbulent eddies \cite{ramanan2019investigation}.
It is thus expected that the flame edge will be conditionally thinner at unstable condition compared to that in stable condition for turbulent flames. Thus, by using flame edge as a marker, distinction can be made across different dynamical states.

The condition mentioned above is necessitated by the fact that although large scale coherent structures modulate the flame edge in an ordered manner, still there is a significant modulation of the flame edge during a cycle of combustion instability. The cycle refers to the time period of the dominant oscillations.
Therefore, any time-averaging over the cycle will result in a thick and meaningless description of flame edge. However, evaluating an ensemble at recurring events will result in overlapping flame edges, which satisfies the objective of distinguishing across different dynamical states. This necessitates us to evaluate flame edge at conditional events, which in the present work are chosen to be local pressure maxima. The conditional event is chosen to occur at time instants where the local pressure exceeds $70\%$ of the peak pressure value of the particular configuration being considered. This value is close to the root mean square value, which is a common metric used in combustion dynamics \cite{sarkar2016dynamic}. Hence, the chosen conditional event is devoid of ambiguity.
Subsequently, the flame images are conditionally ensembled. The flame edge is detected using the standard Canny edge detection algorithm, which is a standard algorithm employed for flame edge detection  \cite{bellows2007flame, zhou2020effects, ahn2015effect}.

The ensemble flame edge computed from the samples around the peak values are shown in Fig.~\ref{3dcsae_500_40_30_validation}(d). We observe that for sample \#179, the conditionally ensembled flame edge is narrow (thin) and it overlaps across the conditional events implying ordered modulation, with the flame seen attached to the swirler.
For samples \#178 and \#180, the ensemble flame edges are comparatively thicker and are seen to occupy a larger space owing to non-periodic modulations which results in a significant scatter of the flame edges evaluated at the conditioned events.
As discussed earlier, the ensemble flame edge is narrow for unstable instants and the flame is always attached to the flame-holder at the unstable conditions. While the sample \#179 exhibits properties of unstable condition, the samples \#178 and \#180 do not, being represented by thicker ensemble flame edges. Similar results are also obtained when sample \#442 is considered.  

We plot the normalized conditioned pressure (time-series based technique) along with computing flame edge (image based technique). If the peak pressure for the entire time-period is $P_{max}$, we include condition in our calculation by considering pressure instants which are greater than $0.7(P_{max})$ based on the explanation provided earlier. For the instants where the unsteady pressure does not meet the conditions, zero values are assigned for clarity. For the selected pressure instants, we normalize by dividing the values with $P_{max}$. As $P_{max}$ can occur in the unstable regime after the transition has happened, normalized conditioned pressure can only be an off-line measure.
The time instants from 3D-CSAE results corresponding to the peaks (sample \#179, sample \#442) are 0.96s and 2.36s respectively (Fig.~\ref{3dcsae_500_40_30_validation}(a)). 
Similar close peaks are observed from the normalized  pressure plot at 1.01s and 2.40s respectively. 
We compare the results with 2D-CSAE which incorrectly identifies an early warning signal at 1.48s (Fig.~\ref{3dcsae_500_40_30_validation}(c)).  

The conditioned pressure instants ahead of the transition can be labeled as intermittent conditions (1.01s and 2.40s) due to their short durations. Around these conditions, we observe thin ensemble flame edge, which denotes unstable condition as stated earlier. We observe that thinning is most prominent along the stream-wise axis of the two-dimensional image. 
Both of these techniques (ensemble flame edge, normalized conditioned pressure) are post-processing techniques. 
Therefore, these techniques cannot act as online detectors, though they are important for validation. Our deep learning based framework has the potential to be employed as an online detector. The successful verification using physics based methods confirms that the 3D-CSAE based detector is correctly picking out the early warning signals.

\section{Discussion}
In this paper, we propose a 3D deep learning model that leverages spatiotemporal data (i.e., flame videos) in detecting the transition from a stable to an unstable regime in combustion systems. Note that the training was performed using completely stable and completely unstable flame videos. The test results demonstrate excellent generalization capability of the trained model as it can identify gradual transition from stable to unstable regimes (though it was not trained with transition data, which is extremely difficult to annotate) as well as precursors to instability present in the stable regime data, all under different operating conditions (one operating condition presented here, more conditions are presented in supplementary material). Therefore, our framework is suitable for real world applications. 

In addition to presenting the data-driven results, we carefully verify the detection of the transition and the precursors by our model using physics-based domain knowledge. Prior work has shown that the flame video modality is more informative in detecting transition and precursors compared to other modalities such as acoustic pressure and chemiluminescence \cite{sarkar2015early1}. We show that our proposed framework performs better than the prior state-of-the-art 2D models such as 2D-CSAE \cite{akintayo2016prognostics} that can both detect transitions as well as provide insights through identifying flame structures relevant to instability. We do not compare performance with other spatial (2D CNN) or spatiotemporal (3D CNN or CNN-LSTM) classification models as we have a different learning objective that involves detecting the transition from stable to unstable regime as well as detecting precursors. 

There has been a recent focus on developing machine learning based online instability detectors for combustion systems. However, such efforts have been purely data-driven exercises using experimental datasets annotated by experts. To the best of our knowledge, this is the first study where deep learning results attempt to offer scientific insights and are verified in details from a physics-based understanding. Therefore, this study can build confidence of the domain experts to use such machine learning models in real systems. By deploying such framework on a real system, early warning signals can be detected effectively which can prevent unwanted incidents (e.g., flame blow out and structural damage to the engines) resulting from transition to instability. The current roadblock for deploying such model on real systems exists primarily on the hardware side as acquisition and processing of high volume flame video data may not be feasible to perform fast enough to mitigate combustion instability using current commercial hardware. Similarly, flame imaging is not a common sensing modality in the engines today. However, with rapid improvement in the hardware sector, we anticipate that a flame video based instability detection framework will be feasible soon and this will be a transformative advancement in terms of energy efficiency and safety for the power generation and transportation industry.

\section*{Supplementary Material}
See the supplementary material for a section with the details of the 2D-CSAE architecture along with the model figure, a section about the results of two other test protocols ($600_{50to35}$ and $500to600_{40}$) with figures and a section with additional 2D-CSAE results along with a figure.

\section*{Authors' Contributions}
S.R.C. and Soumalya S. conceived the combustion experiment(s), V.R. conducted the combustion experiment(s) and collected the data, T.G., A.A., Soumalya S. and Soumik S. have conceived and built the machine learning models, T.G., A.A. and P.K.B. generated the machine learning results, V.R. and T.G. performed the physics-based analysis. T.G., V.R., Soumalya S., S.R.C. and Soumik S. interpreted the results. All authors have contributed in writing and reviewing the manuscript.

\begin{acknowledgments}
This work has been supported in part by the U.S. Air Force Office of Scientific Research under the YIP grant FA9550-17-1-0220. Any opinions, findings and conclusions or recommendations expressed in this publication are those of the authors and do not necessarily reflect the views of the sponsoring agency.
The authors also acknowledge Mr. William Beach who helped in preparing the schematic of the engine shown in Fig. 1.
\end{acknowledgments}

\section*{Data Availability}
The data that support the findings of this study are available from the corresponding author upon reasonable request.


\section*{References}
\bibliography{references}

\clearpage

\section*{Supplementary Material}

\section*{2D Convolutional Selective Autoencoder (2D-CSAE)}
We propose a 3D deep learning model architecture called 3D-CSAE in this paper. We also develop a 2D variant of the architecture, 2D-CSAE as a baseline model. Both models are termed as "selective" because of the training scheme implemented with selective outputs for stable and unstable samples.

\begin{figure*}
\centering
\includegraphics[width=17cm, keepaspectratio]{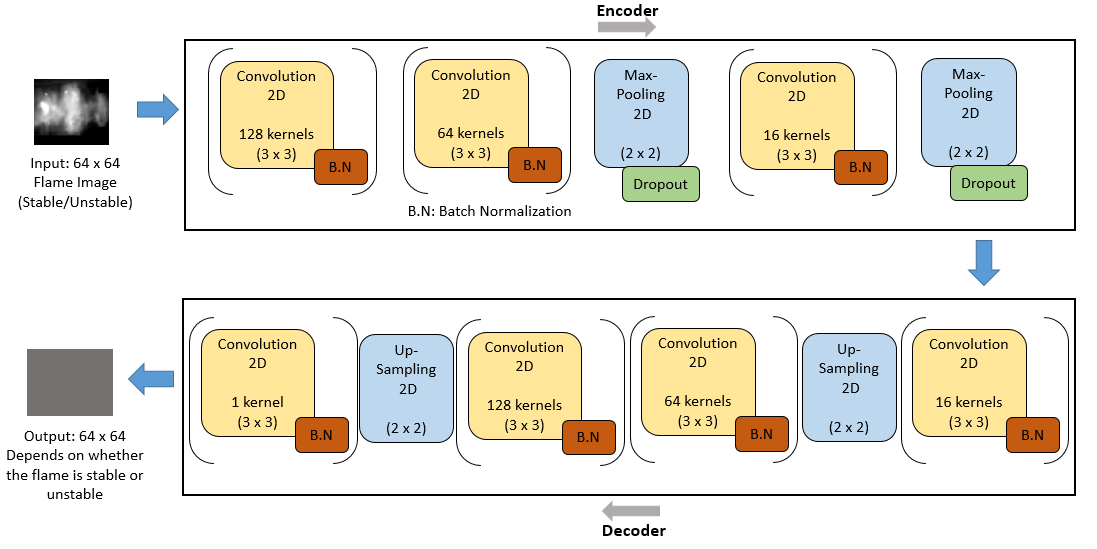}
\caption{The network architecture of the 2D-CSAE Model. Unlike the 3D-CSAE model, for this model both the input and output are two-dimensional. The 2D-CSAE model comprises of 2D convolution, 2D max-pooling and 2D up-sampling layers. }
\label{2dcsae_model} 
\end{figure*}

The architecture for 2D-CSAE is developed by replacing the 3D conv, 3D pooling and 3D up-sampling layers of 3D-CSAE with 2D conv, 2D pooling and 2D up-sampling layers respectively. 2D-CSAE, a 2D version of 3D-CSAE is used for comparative studies. Instead of a volumetric sample, the input to the 2D-CSAE is an image of dimension 64 x 64. After the encoder reduces the image to a lower dimensional representation, the decoder reconstructs the output which is of the same dimension as the input. The 2D-CSAE network architecture is shown in Fig.~\ref{2dcsae_model}.

The 2D conv layers compute the feature maps by extracting spatial information. In the encoder of 2D-CSAE, we have three 2D conv layers and two 2D max-pooling layers. For the 2D conv layers, we use kernel size of 3 x 3 and for the 2D max-pooling layers, the pool size is 2 x 2. The decoder comprises of two 2D up-sampling layers along with four 2D conv layers. 

\section*{3D-CSAE Results for Other Test Protocols}
We present here the 3D-CSAE results for the other two test protocols - $600_{50to35}$ and $500to600_{40}$. 

\begin{figure*}
\centering
\includegraphics[width=17cm, keepaspectratio]{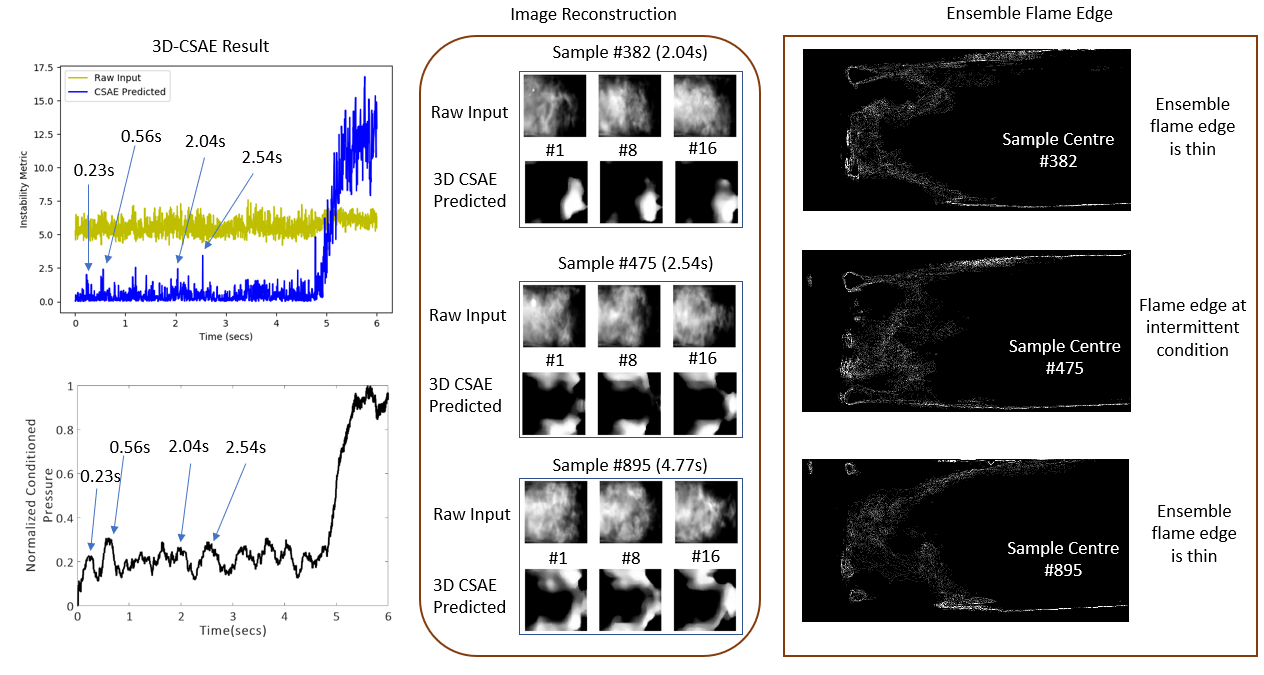}
\caption{The 3D-CSAE detection and validation results for the test protocol $600_{50to35}$. From the instability metric curve, we observe multiple clear peaks before the transition occurs around 5 secs, indicating early-warning signals. Some of the highlighted peaks in the plot are at 0.23s, 0.56s, 2.04s and 2.54s. Output reconstructions, shown for the two volumetric samples corresponding to the peaks at 2.04s and 2.54s, demonstrate more evolution of frame structures with less masking, which implies deviations from the stable condition. Reconstructions for sample \#895 (just before the transition) is shown. Most of the peaks from the normalized conditioned pressure plot are detected by 3D-CSAE. Thin ensemble flame edge structures observed at the intermittent conditions validate that these instants show features of instability for the obtained results.}
\label{3dcsae_600_50_35_results} 
\end{figure*}

In the main paper, the results are shown for the protocol $500_{40to30}$, where the transition is sudden with very few early warning signals, which are correctly captured by the 3D-CSAE. In the test protocol $600_{50to35}$ the transition is also sudden but unlike $500_{40to30}$, here the system is relatively unstable at the beginning with higher number of intermittent conditions (early-warning signals). We observe multiple peaks in the normalized condition pressure plot before the transition occurs and the signals are captured by the 3D-CSAE at similar time instants as shown in Fig.~\ref{3dcsae_600_50_35_results}. We show the reconstructions of two such instants, where less masking is observed in the output reconstructions of the corresponding volumetric samples. The ensemble flame edge structures also validate the results. 

\begin{figure*}
\centering
\includegraphics[width=16cm, keepaspectratio]{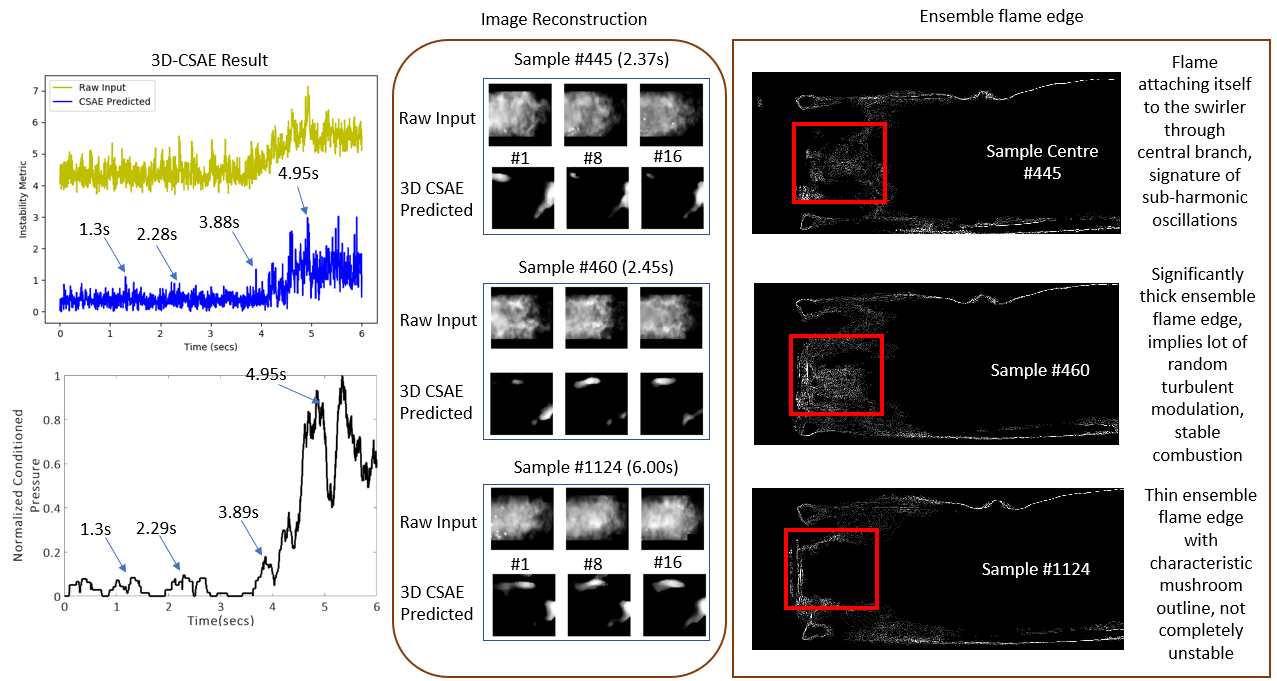}
\caption{The 3D-CSAE detection and validation results for the test protocol $500to600_{40}$. The instability metric curve doesn't show clear peaks in the stable regime and also, the metric values do not increase considerably, unlike what we observe for the protocols $500_{40to30}$ and $600_{50to35}$. This is an indication that proper transition to instability has not occurred. Significant masking is observed even as late as 6s (sample \#1124). The ensemble flame edge results also confirm that the flame is not completely unstable at 6s. Some of the less prominent peaks captured by 3D-CSAE are also seen in the normalized conditioned pressure plot, which shows fluctuating characteristics.}
\label{3dcsae_500_600_40_results} 
\end{figure*}

In the test protocol $500to600_{40}$, proper transition to instability from a stable regime does not occur, the results of which are shown in Fig.~\ref{3dcsae_500_600_40_results}. The 3D-CSAE instability curve does not show clear peaks in the stable regime and even in the later time instants, there is not considerable increase in the values of the instability metric. In the time instant as late as 6s, the frame reconstructions of the volumetric sample are mostly masked with insignificant evolution of the flame structure. Fluctuating characteristics are observed in the normalized conditioned pressure plot. Though the ensemble flame edge features in the earlier time instants indicate stable combustion, the features of the later instants are indicative that the system is not completely unstable. These validate the 3D-CSAE results using physics-based techniques for the protocol $500to600_{40}$. 

\subsection*{2D-CSAE Results}

\begin{figure*}
\centering
\includegraphics[width=16cm, keepaspectratio]{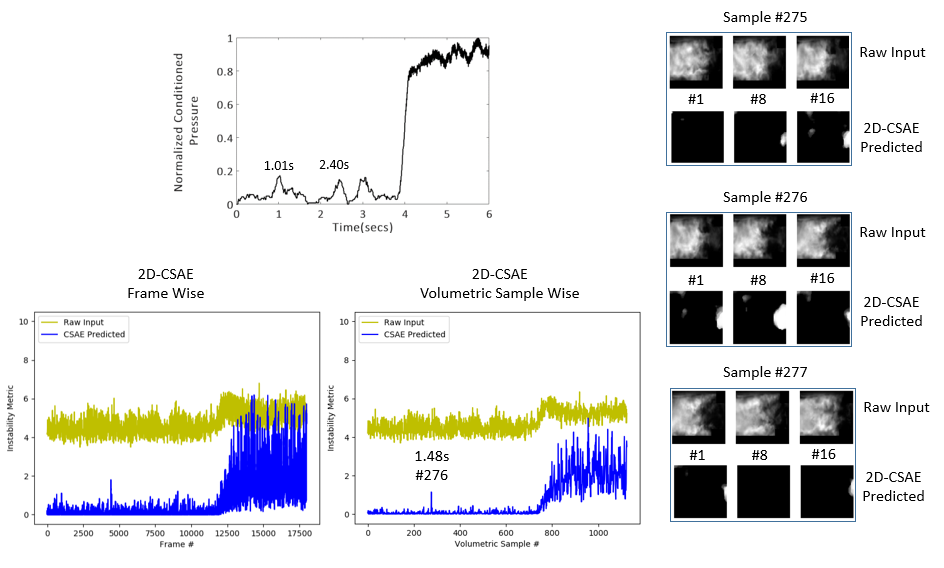}
\caption{The 2D-CSAE results for the test protocol $500_{40to30}$. The instability metric curve is shown for the frame wise output of the 2D-CSAE. For each sample, the volumetric sample wise metric is computed after taking average of the frame wise metrics, considering the same number of frames per sample as in the 3D-CSAE. The peak captured by the 2D-CSAE doesn't match with the results of the physics-based measure (normalized conditioned pressure). With not much significant change in the metric values, the transition is less vividly captured by the 2D-CSAE compared to that by the 3D-CSAE.}
\label{2dcsae_500_40_30_results} 
\end{figure*}

Fig.~\ref{2dcsae_500_40_30_results} shows the 2D-CSAE results for the test protocol $500_{40to30}$. The output reconstruction of 2D-CSAE is frame wise, for which the instability metric curve is shown. Considering the same number of frames per sample (16) (as it was for the 3D-CSAE) and taking the average for all frames, the volumetric sample wise metric is computed. In the stable regime, the peak captured by the 2D-CSAE is at time instant 1.48s, which does not show similarity with the results based on the normalized conditioned pressure. Also, as compared to 3D-CSAE, the transition is less clear (not a significant increase in the metric values) in the 2D-CSAE results.  

\end{document}